\documentclass{article}
\usepackage{spconf,amsmath,graphicx,hyperref}
\usepackage{multirow} 
\usepackage{amsmath}
\usepackage{amssymb}
\usepackage{booktabs} 
\usepackage{multirow}
\usepackage[export]{adjustbox}
\usepackage{algpseudocode}
\usepackage{algorithm}
\usepackage{float}
\usepackage{stfloats}
\usepackage{dsfont}
\usepackage{rotating}
\usepackage{array}
\usepackage{xcolor}
\title{Aligning What You Separate: Denoised Patch Mixing for Source-Free Domain Adaptation in Medical Image Segmentation}
\name{%
  \begin{tabular}[t]{@{}c@{}} % centered column, no extra side padding
    Quang-Khai Bui-Tran$^{2,3}$* \qquad 
    Thanh-Huy Nguyen$^1$* \qquad 
    Hoang-Thien Nguyen$^2$* \qquad 
    Ba-Thinh Lam$^2$ 
    \\ 
    Nguyen Lan Vi Vu$^4$ \qquad 
    Phat K. Huynh$^2$ \qquad 
    Ulas Bagci$^5$ \qquad 
    Min Xu$^1$**\thanks{* Equal Contribution} \thanks{** Corresponding Author: mxu1@cs.cmu.edu}
  \end{tabular}%
}

\address{
$^1$Carnegie Mellon University, PA, USA \\
$^2$PASSIO Lab, North Carolina A\&T State University, NC, USA \\
$^3$Ho Chi Minh University of Science, Vietnam \\
$^4$Ho Chi Minh University of Technology, Vietnam \\
$^5$Northwestern University, IL, USA
}
\begin{document}
%\ninept
%
\maketitle

\begin{abstract}
Source-Free Domain Adaptation (SFDA) is emerging as a compelling solution for medical image segmentation under privacy constraints, yet current approaches often ignore sample difficulty and struggle with noisy supervision under domain shift. We present a new SFDA framework that leverages \textbf{Hard Sample Selection} and \textbf{Denoised Patch Mixing (DPM)} to progressively align target distributions. First, unlabeled images are partitioned into reliable and unreliable subsets through entropy-similarity analysis, allowing adaptation to start from easy samples and gradually incorporate harder ones. Next, pseudo-labels are refined via Monte Carlo-based denoising masks, which suppress unreliable pixels and stabilize training. Finally, intra- and inter-domain objectives mix patches between subsets, transferring reliable semantics while mitigating noise. Experiments on benchmark datasets show consistent gains over prior SFDA and UDA methods, delivering more accurate boundary delineation and achieving state-of-the-art Dice and ASSD scores. Our study highlights the importance of progressive adaptation and denoised supervision for robust segmentation under domain shift.
\end{abstract}


\section{Introduction}

Deep neural networks have recently made significant contributions to related fields, particularly medical image analysis \cite{FDCL, fetal-bcp}. However, they are highly vulnerable to domain shifts between training and testing datasets, which causes them to perform poorly in real-world scenarios \cite{DPL}. 
Source-free domain adaptation (SFDA) has emerged as a promising paradigm to address domain shift by eliminating reliance on source data during adaptation and leveraging only unlabeled target data \cite{li2024comprehensive, BEAL, DPL, CPR, PLPB, yaacovi2025source}. BEAL \cite{BEAL} mitigates the uncertainty in soft boundary regions via adversarial learning. DPL \cite{DPL} introduces a denoising strategy to enhance self-training, thereby providing more discriminative and less noisy supervision. CPR \cite{CPR} identifies context-inconsistent predictions and proposes context-aware pseudo-label refinement to improve adaptation. PLPB \cite{PLPB} designs a pseudo-boundary loss to exploit edge information from both domains, enabling more accurate predictions in boundary regions. Recently, Yaacovi et al. \cite{yaacovi2025source} leveraged the foundation model SAM \cite{SAM} to evaluate pseudo-label quality and mitigate noisy supervision during target model training.

\begin{figure}[t]
    \centering
    \includegraphics[width=.9\linewidth]{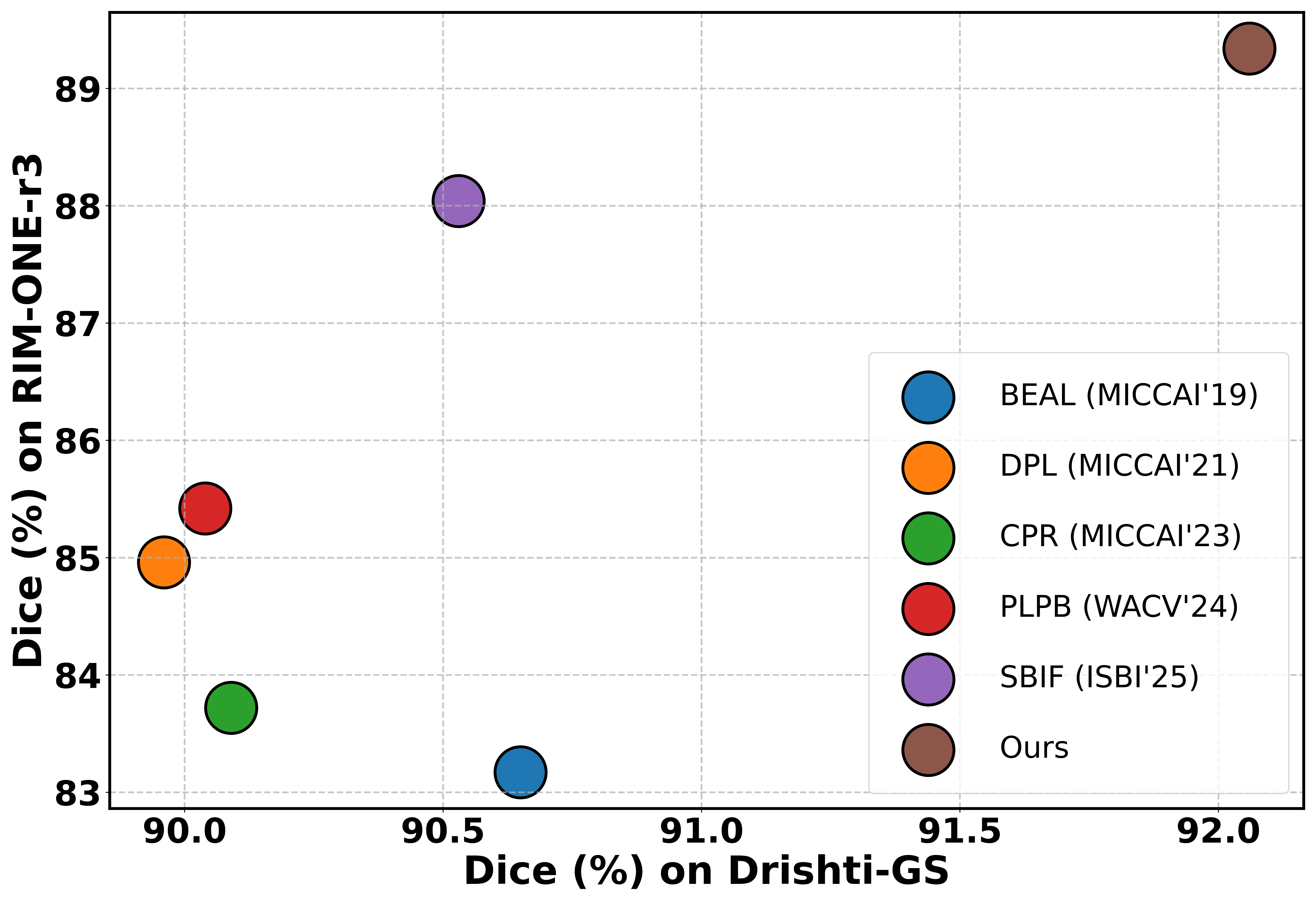}
    \caption{Performance comparison of different methods under the same backbone. The proposed framework significantly boosts Dice score compared to previous works.}
    \label{fig:dice_metrics}
\end{figure}

In medical image segmentation, most existing approaches adapt to the entire target domain at once, overlooking the varying difficulty of samples. Early inclusion of hard samples often introduces noisy pseudo-labels, which destabilize self-learning. We argue that effective adaptation requires a progressive strategy: starting with easy samples that incur minimal errors to capture the domain’s distribution and context, and then gradually incorporating harder samples. To prevent noise propagation during this process, we introduce filtering mechanisms that selectively suppress unreliable pixel-level predictions, enabling stable and robust adaptation across the full target domain. 

Building on this insight, we propose a framework that exploits only unlabeled target data for medical segmentation. Specifically, we partition the target domain into reliable and unreliable subsets via a \textit{Hard Sample Selection} procedure, combining entropy- and similarity-based criteria for accurate separation. Pseudo-labels are then generated within a teacher-student framework~\cite{mean-teacher}, refined by a Monte Carlo (MC)-based noisy filter mask. This forms the core of our \textit{Denoised Patch Mixing} (DPM) technique,  which enhances pseudo-label reliability during the mixing process. To mitigate distribution mismatch, we further introduce a dual alignment strategy: \textit{Intra-Domain Alignment} within reliable samples and \textit{Inter-Domain Alignment} between reliable and unreliable subsets. Extensive experiments on fundus image segmentation benchmarks demonstrate the effectiveness and promise of our method. As shown in Fig. \ref{fig:dice_metrics}, our method achieves superior performance.
\section{Methodology}
\begin{figure*}
    \centering
    \includegraphics[width=.9\linewidth]{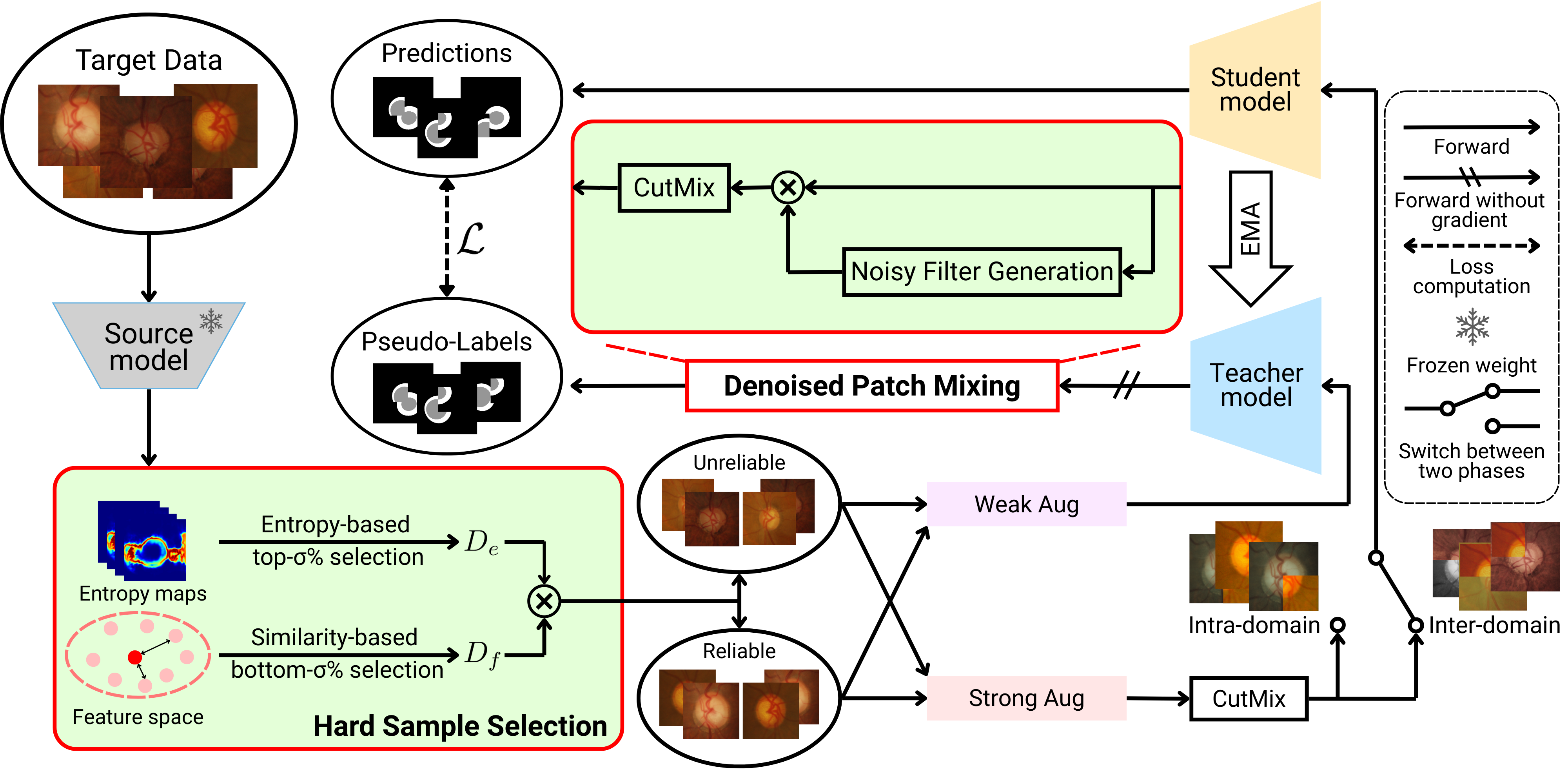}
    \caption{Overview of our proposed framework for source-free domain adaptive medical image segmentation.}
    \label{fig:main_figure}
\end{figure*}

In the Source-Free Domain Adaptation (SFDA) setting, with the provided source model $f_{s}$ trained on the labeled source dataset $\mathcal{D_{S}}$
% $\mathcal{D_{S}} = \{(x_s^i, y_s^i)\}$
, the goal is to adapt the target model $f_{t}$ using only the unlabeled target dataset $\mathcal{D_{T}} = \{x_t^i\}$ with the given the source model $f_{s}$. Our framework is illustrated in Fig. \ref{fig:main_figure}.

\subsection{Hard Sample Selection}
Under domain shift, target data often contains hard samples making self-training with pseudo-labels challenging. Including these hard samples that generate incorrect pseudo-labels in training can lead to error accumulation. To address this, we propose constructing two subsets using a hard-sample selection mechanism.  

\textbf{Entropy-based selection.} Some hard samples exhibit high entropy where the model is uncertain about its predictions due to domain discrepancy. Given a trained source model, we perform inference on the target dataset and compute the entropy of each sample as  
\begin{equation}
e(x_t) = - \sum_{v} p_v(x_t) \log\big(p_v(x_t)\big),
\end{equation}
where $p_v(x_t)$ is the output probability at pixel $v$ of sample $x_t$. We then select the top-$\sigma\%$ high-entropy samples:
\begin{equation}
D_e = \{\, x_t^i \in D_T \;\mid\; i \in \text{Top-}\sigma\%(e(x_t)) \,\}.
\end{equation}

\textbf{Feature similarity-based selection.}  
The source model can sometimes produce over-confident predictions and miss these hard sample. Also the raw predictions may be biased toward the source distribution. Since entropy alone may not be reliable, we additionally incorporate feature-based similarity. Specifically,
\begin{equation}
z_i = E(x_t^i), \quad 
z_p = \frac{1}{n} \sum_{i=1}^{n} E(x_t^i),
\end{equation}
where $E(\cdot)$ denotes the backbone encoder, $z_i$ is the feature of the $i$-th sample, and $z_p$ is the centroid feature computed from the source model. We then compute the cosine similarity and select the $\sigma\%$ samples with the lowest similarity:
\begin{equation}
D_f = \{\, x_t^i \in D_T \;\mid\; i \in \text{Bottom-}\sigma\%(\cos(z_i, z_p)) \,\}.
\end{equation}
\textbf{Final subset construction.}  
The unreliable subset is defined as the intersection of the two criteria, combining both feature-level similarity and prediction uncertainty:
\begin{equation}
D_u = D_e \cap D_f,
\end{equation}
while the remaining dataset is considered reliable for adaptation. These reliable samples better reflect the global distribution of $D_T$ due to their higher similarity with the feature prototype, with $D_T$ denoting the full target dataset:
\begin{equation}
D_r = D_T \setminus D_u.
\end{equation}

\subsection{Pseudo-label and Noisy Filter Generation}
Unlike image classification, which makes predictions at the image level, image segmentation produces predictions at the pixel level. Under domain shift, these pixel-wise predictions are often uncertain and may contain noisy or incorrect pixels even with the easy-to-adapt samples. To mitigate this, we use the teacher model to generate pseudo-labels and the corresponding noisy-pixel filter mask. 

For each target image $x_t$, we compute $K$ stochastic predictions $p_{k} = f^T(x_t)$ for $k = 1, \dots, K$, and derive the mean $p = \text{avg}(p_1, \dots, p_K)$ and standard deviation $\tau = \text{std}(p_1, \dots, p_K)$. Using a confidence threshold $\gamma$, the pseudo-label is defined as $\hat{y} = \mathds{1}[p \geq \gamma]$ with $\eta$ being a predefined uncertainty threshold and $v$ denotes the pixel index.

For each class $c \in \{0,1\}$, the prototype is computed using only low-uncertainty features with $z$ denotes the feature map:
\begin{equation}
\mu^{c} = 
\frac{\sum_{v} z_v \cdot \mathds{1}[\hat{y}_v = c] \cdot \mathds{1}[\tau_v < \eta] \cdot p_v}
     {\sum_{v} \mathds{1}[\hat{y}_v = c] \cdot \mathds{1}[\tau_v < \eta] \cdot p_v}.
\end{equation}

The distance of each pixel to its class prototype is calculated, and the denoising mask is defined as:
\begin{equation}
d_v^{c} = \left\lVert z_v - \mu^{c} \right\rVert_2,
\end{equation}
\begin{equation}
m_v = \mathds{1}[\hat{y}_v = 1] \cdot \mathds{1}[d_v^{1} < d_v^{0}] 
    + \mathds{1}[\hat{y}_v = 0] \cdot \mathds{1}[d_v^{1} > d_v^{0}].
\end{equation}

\subsection{Intra-domain and Inter-domain Alignment}
To fully exploit the target domain, we adopt a two-stage training strategy: first, training on the reliable subset $D_r$, then gradually incorporating the unreliable subset $D_u$. However, distributional variance may arise both within $D_r$ and between $D_r$ and $D_u$, and this variance may further increase under the two-stage strategy. To mitigate this, we propose Denoised Patch Mixing (DPM), which leverages pseudo-labels $\hat{y}$ and denoising masks $m$ from the teacher model $f^T$. DPM provides semantic context by mixing image regions while suppressing noisy labels via denoising masks, minimizing error transfer during mixing. This enables the model to benefit from mixed supervision that aligns distributions intra-domain (within $D_r$) and inter-domain (between $D_r$ and $D_u$). Consequently, DPM reduces noise sensitivity, improves pseudo-label quality, and enhances robustness.

Given two images with strong augmentations, their pseudo-labels and denoising masks $(x_1,\hat{y}_1,m_1), (x_2,\hat{y}_2,m_2)$, and a binary DPM region $M \in \{0,1\}^{H \times W}$, we form:

\begin{align}
\tilde{x} &= M \odot x_1 + (1-M)\odot x_2, \\
\tilde{y} &= M \odot \hat{y}_1 + (1-M)\odot \hat{y}_2, \\
\tilde{m} &= M \odot m_1 + (1-M)\odot m_2,
\end{align}
where $\odot$ denotes element-wise multiplication. The mixed mask $\tilde m$ indicates reliable pixels in the mixed sample.

Let the student's prediction on $\tilde x$ be $p^S(\tilde x)$, representing pixel-wise class probabilities. We define the pixel-wise binary cross-entropy as usual and introduce a mask-normalized loss that only considers pixels selected by the binary mask $m$:
\begin{equation}
\mathcal{L}_{\mathrm{BCE}}(p, y; m) = 
\frac{\sum_v m_v \, \ell_{\mathrm{BCE}}(p_v, y_v)}
{\sum_v m_v}.
\end{equation}

\noindent\textbf{Intra-domain alignment.} 
For intra-domain alignment, we draw $(x_1,\hat{y}_1,m_1), (x_2,\hat{y}_2,m_2)\sim D_r$ and construct $(\tilde x,\tilde y,\tilde m)$ using Denoised Patch Mixing. The intra-domain loss is:
\begin{equation}
\mathcal{L}_{\mathrm{intra}}
= \mathbb{E}_{(x_1,x_2)\sim D_r}
\big[\,\mathcal{L}_{\mathrm{BCE}}\big(p^S(\tilde x),\;\tilde y;\;\tilde m\big)\,\big].
\end{equation}
This encourages the model to learn consistent semantics within $D_r$, improving robustness and ensuring smooth transitions for the next stage.

\noindent\textbf{Inter-domain alignment.} 
For inter-domain alignment, we mix one reliable and one unreliable sample: $(x_r,\hat{y}_r,m_r)\sim D_r,\; (x_u,\hat{y}_u,m_u)\sim D_u$ to form $(\tilde x,\tilde y,\tilde m)$ and define:
\begin{equation}
\mathcal{L}_{\mathrm{inter}}
= \mathbb{E}_{(x_r,x_u)\sim D_r\times D_u}
\big[\,\mathcal{L}_{\mathrm{BCE}}\big(p^S(\tilde x),\;\tilde y;\;\tilde m\big)\,\big].
\end{equation}
The cross-set Denoised Patch Mixing transfers reliable and comprehensive semantics from $D_r$ into the unreliable regions of $D_u$, effectively reducing the difficulty of learning from $D_u$ while enabling adaptation to the full target distribution without leaving out any samples.

Afterward, the teacher parameters $\theta_t^{(k+1)}$ at iteration $k+1$ are updated with smoothing coefficient $\alpha$ as
\[
\theta_t^{(k+1)} \gets \alpha\theta_t^{(k)}+(1-\alpha)\theta_s^{(k)}.
\]

\begin{table}[t]
\centering
\setlength{\tabcolsep}{0.8mm} % reduce column separation
\renewcommand{\arraystretch}{1}
\begin{tabular}{|l|c|c|c|c|c|}
\hline
\multirow{2}{*}{\textbf{Method}} & \multirow{2}{*}{\textbf{S-F}} &
\multicolumn{2}{c|}{\textbf{Optic Disc Seg.}} & 
\multicolumn{2}{c|}{\textbf{Optic Cup Seg.}} \\
\cline{3-6}
& & Dice $\uparrow$ & ASSD $\downarrow$ & Dice $\uparrow$ & ASSD $\downarrow$ \\
\hline
\multicolumn{6}{|c|}{\textbf{Drishti-GS}} \\
\hline
Source only &  & 94.04  & 7.47 & 80.22 & 13.47 \\
Target only &  & 97.40 & 3.58 & 90.10 & 9.50 \\
\hline
BEAL\cite{BEAL}  & $\times$ & 96.12  & 4.48 & \underline{85.18} & \underline{9.66}\\
DPL \cite{DPL}  & \checkmark & 96.39  & 4.08 & 83.53 & 11.39 \\
CPR \cite{CPR}  & \checkmark & 96.36  & 4.09 & 83.81 & 10.96 \\
PLPB\cite{PLPB} & \checkmark & 96.51  & 4.01 & 83.56 & 11.11 \\
SBIF  \cite{yaacovi2025source}  & \checkmark & \underline{96.59} & \underline{3.92} & 84.47 & 10.21 \\
\textbf{Ours} & \checkmark & \textbf{96.77} & \textbf{3.64} & \textbf{87.34} & \textbf{8.25} \\
\hline
\multicolumn{6}{|c|}{\textbf{RIM-ONE-r3}} \\
\hline
Source only &  & 83.02  & 23.36 & 73.10 & 13.87 \\
Target only &  & 95.74 & 6.05 & 83.97 & 5.38 \\
\hline
BEAL \cite{BEAL}   & $\times$ & 90.28  & 8.95 & 76.06 & \underline{7.19} \\
DPL  \cite{DPL} & \checkmark & 90.13  & 9.43 & 79.78 & 9.01 \\
CPR \cite{CPR}  & \checkmark & 92.39  & 6.86 & 75.04 & 10.43 \\
PLPB \cite{PLPB} & \checkmark & 92.89  & 6.52 & 77.94 & 10.07 \\
SBIF \cite{yaacovi2025source}  & \checkmark & \underline{93.81}  & \underline{5.58} & \underline{82.26} & 7.79 \\
\textbf{Ours} & \checkmark & \textbf{95.14} & \textbf{4.25} & \textbf{83.53} & \textbf{6.70} \\
\hline
\end{tabular}
\caption{Quantitative results with different methods.}
\label{tab:main_result}
\vspace{-3mm}
\end{table}

\begin{figure}[!ht]
    \centering
    \renewcommand{\arraystretch}{1.1}
    \setlength{\tabcolsep}{1pt}
    \resizebox{\linewidth}{!}{%
    \begin{tabular}{rcccccc}
        % Drishti-GS sample
        \raisebox{0.95\height}{\rotatebox[origin=c]{90}{\scriptsize Drishti-GS}} &
        \includegraphics[width=0.13\linewidth]{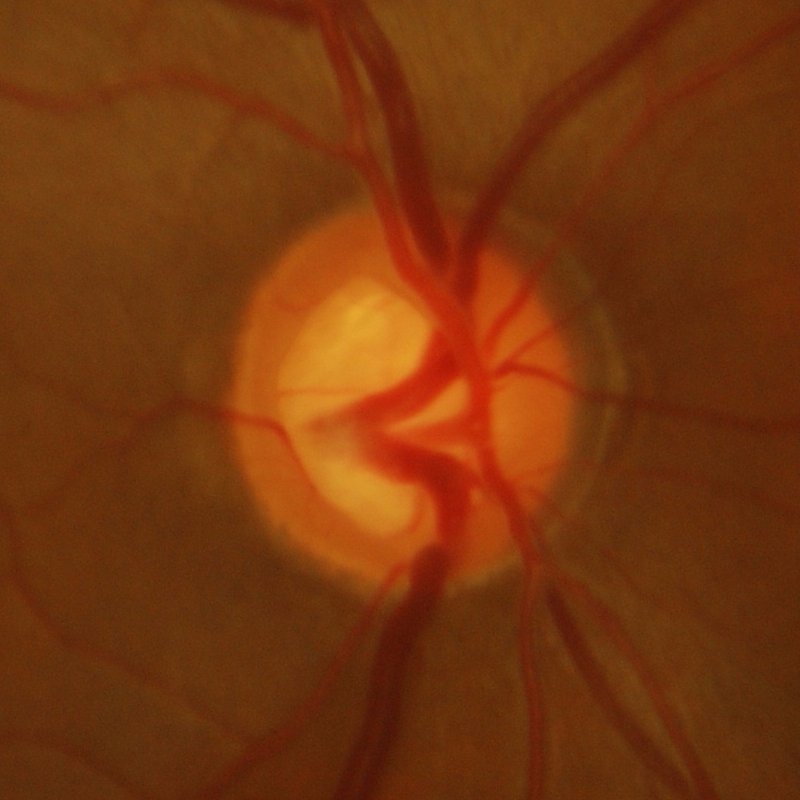} & 
        \includegraphics[width=0.13\linewidth]{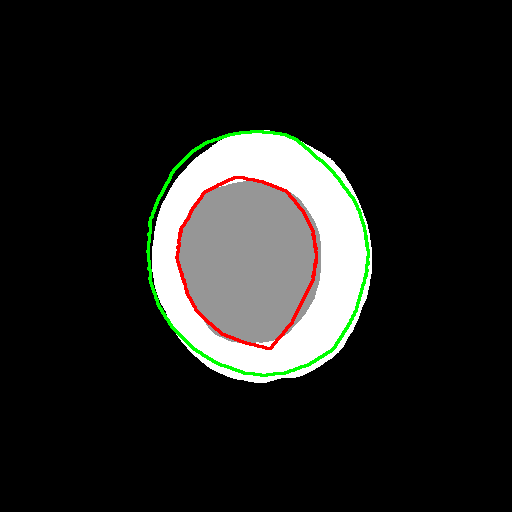} & 
        \includegraphics[width=0.13\linewidth]{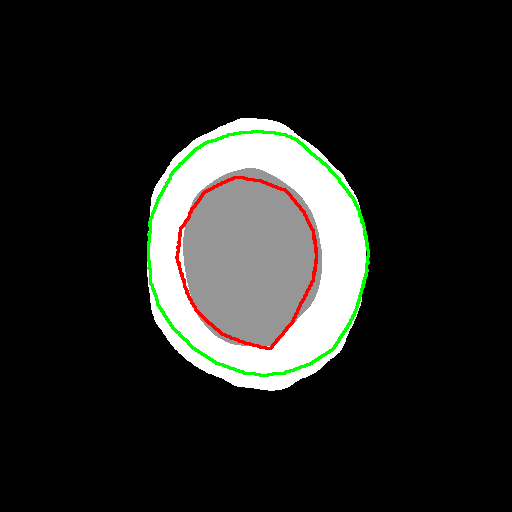} & 
        \includegraphics[width=0.13\linewidth]{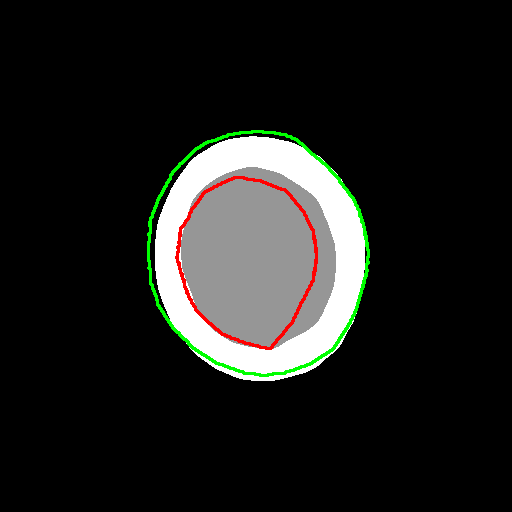} & 
        \includegraphics[width=0.13\linewidth]{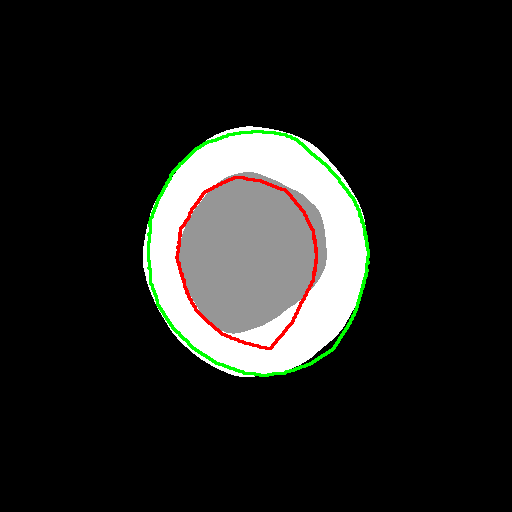} & 
        \includegraphics[width=0.13\linewidth]{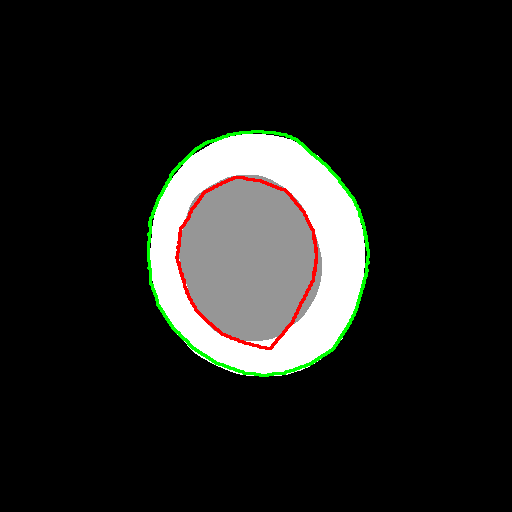} \\
        
        % RIM-ONE sample
        \raisebox{0.85\height}{\rotatebox[origin=c]{90}{\scriptsize RIM-ONE}} & 
        \includegraphics[width=0.13\linewidth]{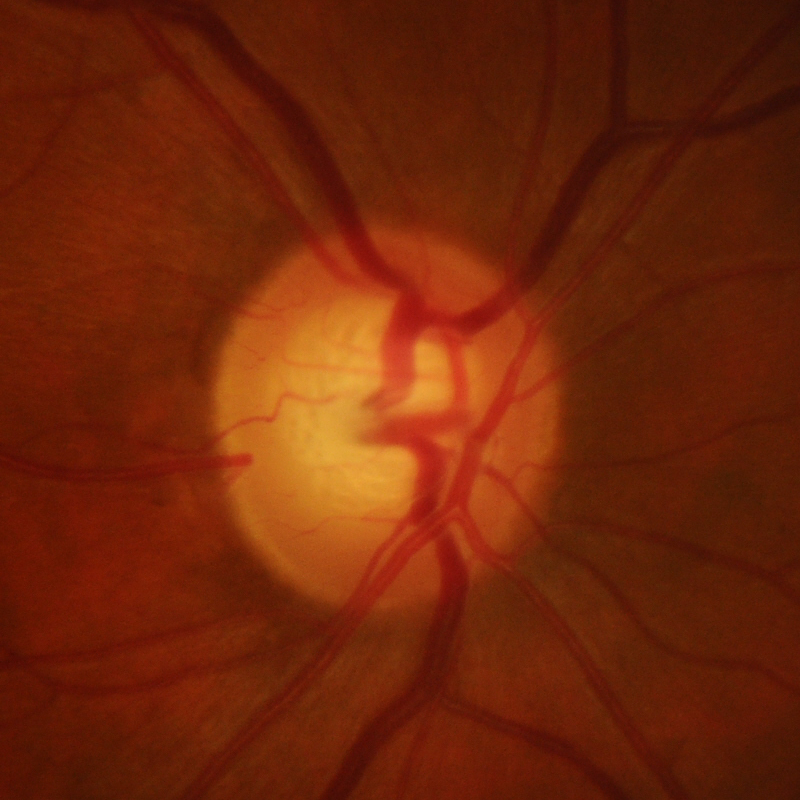} & 
        \includegraphics[width=0.13\linewidth]{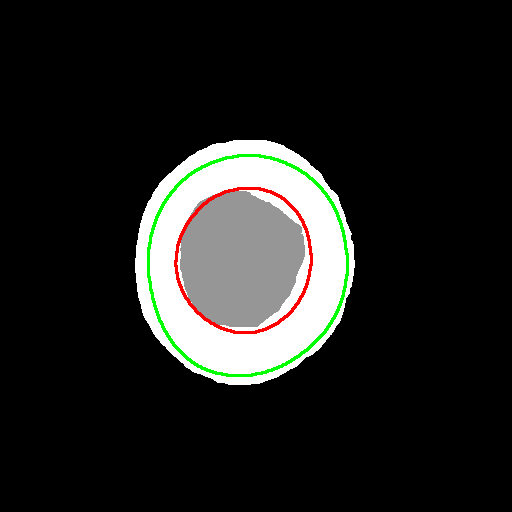} & 
        \includegraphics[width=0.13\linewidth]{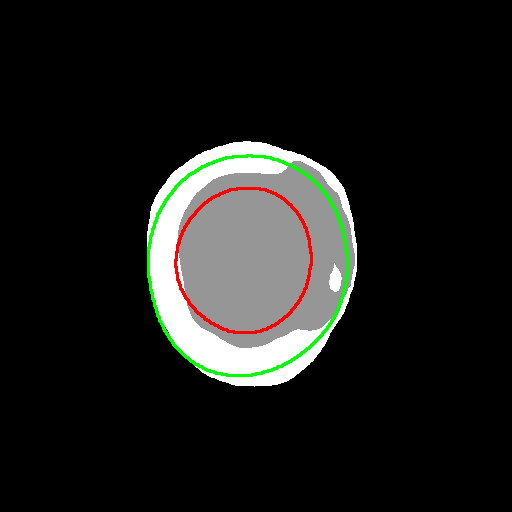} & 
        \includegraphics[width=0.13\linewidth]{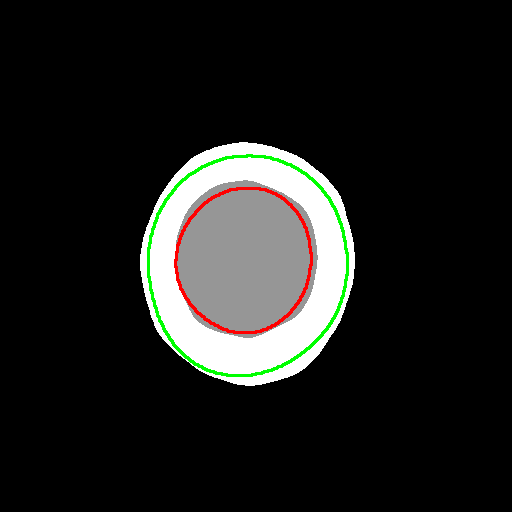} & 
        \includegraphics[width=0.13\linewidth]{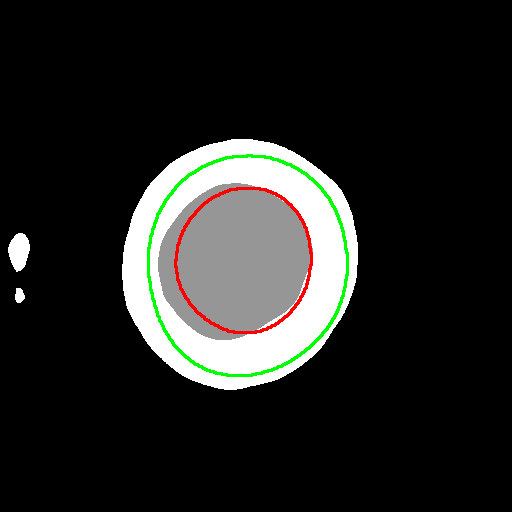} & 
        \includegraphics[width=0.13\linewidth]{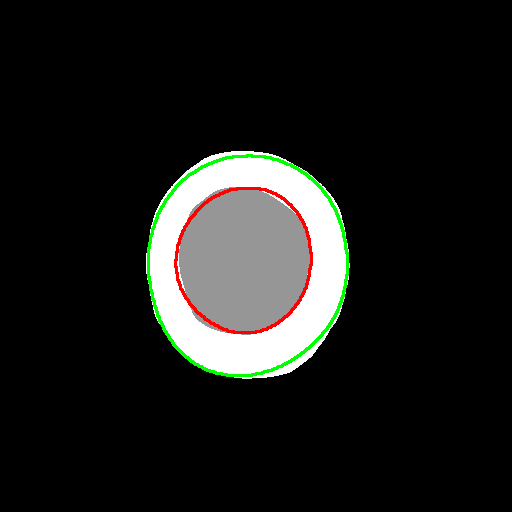} \\

        & \scriptsize Image 
        & \scriptsize BEAL 
        & \scriptsize DPL 
        & \scriptsize CPR 
        & \scriptsize PLPB 
        & \scriptsize \textbf{Ours} \\
    \end{tabular}
    }
    \caption{Qualitative comparisons of different methods.}
    \label{fig:qualitative_results}
\end{figure}

\section{Experiments}

\subsection{Implementation Details}
We use DeepLabV3+ with a MobileNetV2 backbone, setting the confidence threshold to $\gamma=0.75$. The source model is trained for 200 epochs with Adam (lr=$1\times10^{-3}$), and source-free adaptation runs for 10 epochs each in intra- and inter-domain stages (lr=$5\times10^{-4}$). Teacher and student are initialized from the source model, with the teacher updated via EMA ($\alpha=0.99$). We set $\eta=0.05$ with 10 stochastic passes and define the unreliable set as 10\% of the target data. Strong augmentations (contrast, erasing, Gaussian noise) are applied. Experiments use PyTorch on an NVIDIA RTX 3060 GPU. Following prior work, REFUGE~\cite{orlando2020refuge} is the source domain, and RIM-ONE-r3~\cite{fumero2011rim} and Drishti-GS~\cite{sivaswamy2015comprehensive} are targets. Dice and ASSD are used for evaluation.

\subsection{Comparisons and Discussions}
We evaluate our approach with UDA methods like BEAL~\cite{BEAL} and leading SFDA techniques, including DPL~\cite{DPL}, CPR~\cite{CPR}, PLPB~\cite{PLPB}, and SBIF~\cite{yaacovi2025source}, on benchmark fundus datasets. For reference, we report a fully supervised \textit{Target Only} upper bound and a \textit{Source Only} baseline without target access. As shown in Table~\ref{tab:main_result}, our framework achieves state-of-the-art results across all metrics, outperforming even source-aware methods such as BEAL. These gains stem from the denoising filter, which produces cleaner pseudo-labels, and from progressively partitioning target data into easy and hard subsets, enabling effective intra- and inter-domain learning.

Figure~\ref{fig:qualitative_results} shows qualitative comparisons. Our predictions yield sharper boundaries and closely match the ground-truth contours (\textcolor{red}{red} and \textcolor{green}{green}), whereas competing approaches exhibit overlaps and boundary errors.

\textbf{Unreliable size.} Table~\ref{tab:unreliable_size} shows the impact of varying the proportion of target data in the unreliable set. At 1\%, filtering is minimal and yields little gain. Performance improves as the ratio increases, peaking at 10\%. Beyond this point, the unreliable set becomes too large, limiting data for intra-domain alignment and hindering effective inter-domain adaptation.

\textbf{Components.} Table~\ref{tab:components} analyzes the contribution of each module. The baseline is a teacher--student model with filter denoising. Adding Denoised Patch Mixing (DPM) at an early stage degrades performance, as hard samples introduce noisy pseudo-labels. Using only the reliable set offers limited gains, as it neither adapts to the full distribution nor exploits all data. With DPM, the model captures richer context and improves robustness, and combining it with the unreliable set yields the best overall results.

\begin{table}[t]
\centering
\resizebox{\columnwidth}{!}{
\begin{tabular}{|c|c|c|}
\hline
\textbf{Unreliable Set ($\sigma$)} & \textbf{Drishti-GS} & \textbf{RIM-ONE-r3} \\
\hline
$1\%$   & 90.98 & 88.15 \\
$5\%$   & 91.62 & 88.59 \\
\textbf{$10\%$}  & \textbf{91.88} & \textbf{89.08} \\
$15\%$  & 91.78 & 88.95 \\
$25\%$  & 91.41 & 88.73 \\
\hline
\end{tabular}
}
\vspace{-1mm}
\caption{Ablation on different unreliable set ratios $\sigma$.}
\label{tab:unreliable_size}
\vspace{-2mm}
\end{table}

\begin{table}[t]
    \centering
    \setlength{\tabcolsep}{3pt}
    \renewcommand{\arraystretch}{1.1}
    \begin{tabular}{l|cc|cc}
        \toprule
        \multirow{2}{*}{\textbf{Configuration}} & \multicolumn{2}{c|}{\textbf{Drishti-GS}} & \multicolumn{2}{c}{\textbf{RIM-ONE-r3}} \\
        & Dice$\uparrow$ & ASSD$\downarrow$ & Dice$\uparrow$ & ASSD$\downarrow$ \\
        \midrule
        Baseline & 90.71 & 7.08 & 88.47 & 6.22 \\
        + DPM & 90.65 & 7.13 & 88.36 & 5.72 \\
        + Reliable set & 90.74 & 6.99 & 88.43 & 6.18 \\
            \hspace{4mm} + DPM & 91.36 & 6.52 & 88.96 & 5.65 \\
        + Full modules & \textbf{92.06} & \textbf{5.95} & \textbf{89.34} & \textbf{5.48} \\
        \bottomrule
    \end{tabular}
    \vspace{-1mm}
    \caption{Ablation study of different proposed components.}
    \label{tab:components}
    \vspace{-3mm}
\end{table}

\section{Conclusion}
In this work, we propose a framework for SFDA in fundus image segmentation. A Hard Sample Selection module divides target data into reliable and unreliable subsets, followed by a Denoised Patch Mixing strategy that refines pseudo-labels and aligns their distributions. Experimental results show the effectiveness of our model in bridging distributions between subsets and producing accurate boundary predictions.

% -------------------------------------------------------------------------
\bibliographystyle{IEEEbib}
\bibliography{strings,refs}

\begin{thebibliography}{10}

\bibitem{FDCL}
Thanh-Huy Nguyen, Nguyen Lan~Vi Vu, Hoang-Thien Nguyen, Quang-Vinh Dinh, Xingjian Li, and Min Xu,
\newblock ``Semi-supervised histopathology image segmentation with feature diversified collaborative learning,''
\newblock in {\em Proceedings of The First AAAI Bridge Program on AI for Medicine and Healthcare}. 25 Feb 2025, vol. 281 of {\em Proceedings of Machine Learning Research}, pp. 165--172, PMLR.

\bibitem{fetal-bcp}
Ha-Hieu Pham, Le~Tran Quoc~Khanh, Hoang-Thien Nguyen, Nguyen~Lan Vi~Vu, Quang-Vinh Dinh, Thanh-Huy Nguyen, Xingjian Li, and Min Xu,
\newblock ``Fetal-bcp: Addressing empirical distribution gap in semi-supervised fetal ultrasound segmentation,''
\newblock in {\em 2025 IEEE 22nd International Symposium on Biomedical Imaging (ISBI)}, 2025, pp. 1--4.

\bibitem{DPL}
Cheng Chen, Quande Liu, Yueming Jin, Qi~Dou, and Pheng-Ann Heng,
\newblock ``Source-free domain adaptive fundus image segmentation with denoised pseudo-labeling,''
\newblock in {\em Medical Image Computing and Computer Assisted Intervention--MICCAI 2021: 24th International Conference, Strasbourg, France, September 27--October 1, 2021, Proceedings, Part V 24}. Springer, 2021, pp. 225--235.

\bibitem{li2024comprehensive}
Jingjing Li, Zhiqi Yu, Zhekai Du, Lei Zhu, and Heng~Tao Shen,
\newblock ``A comprehensive survey on source-free domain adaptation,''
\newblock {\em IEEE Transactions on Pattern Analysis and Machine Intelligence}, vol. 46, no. 8, pp. 5743--5762, 2024.

\bibitem{BEAL}
Shujun Wang, Lequan Yu, Kang Li, Xin Yang, Chi-Wing Fu, and Pheng-Ann Heng,
\newblock ``Boundary and entropy-driven adversarial learning for fundus image segmentation,''
\newblock in {\em Medical Image Computing and Computer Assisted Intervention--MICCAI 2019: 22nd International Conference, Shenzhen, China, October 13--17, 2019, Proceedings, Part I 22}. Springer, 2019, pp. 102--110.

\bibitem{CPR}
Zheang Huai, Xinpeng Ding, Yi~Li, and Xiaomeng Li,
\newblock ``Context-aware pseudo-label refinement for source-free domain adaptive fundus image segmentation,''
\newblock in {\em International Conference on Medical Image Computing and Computer-Assisted Intervention}. Springer, 2023, pp. 618--628.

\bibitem{PLPB}
Lingrui Li, Yanfeng Zhou, and Ge~Yang,
\newblock ``Robust source-free domain adaptation for fundus image segmentation,''
\newblock in {\em Proceedings of the IEEE/CVF Winter Conference on Applications of Computer Vision}, 2024, pp. 7840--7849.

\bibitem{yaacovi2025source}
Bar Yaacovi and Jacob Goldberger,
\newblock ``Source free domain adaptation with pseudo-labeling quality assessed by sam in fundus image segmentation,''
\newblock in {\em 2025 IEEE 22nd International Symposium on Biomedical Imaging (ISBI)}. IEEE, 2025, pp. 1--5.

\bibitem{SAM}
Alexander Kirillov, Eric Mintun, Nikhila Ravi, Hanzi Mao, Chloe Rolland, Laura Gustafson, Tete Xiao, Spencer Whitehead, Alexander~C Berg, Wan-Yen Lo, et~al.,
\newblock ``Segment anything,''
\newblock in {\em Proceedings of the IEEE/CVF international conference on computer vision}, 2023, pp. 4015--4026.

\bibitem{mean-teacher}
Antti Tarvainen and Harri Valpola,
\newblock ``Mean teachers are better role models: Weight-averaged consistency targets improve semi-supervised deep learning results,''
\newblock {\em Advances in neural information processing systems}, vol. 30, 2017.

\bibitem{orlando2020refuge}
Jos{\'e}~Ignacio Orlando, Huazhu Fu, Jo{\~a}o~Barbosa Breda, Karel Van~Keer, Deepti~R Bathula, Andr{\'e}s Diaz-Pinto, Ruogu Fang, Pheng-Ann Heng, Jeyoung Kim, JoonHo Lee, et~al.,
\newblock ``Refuge challenge: A unified framework for evaluating automated methods for glaucoma assessment from fundus photographs,''
\newblock {\em Medical image analysis}, vol. 59, pp. 101570, 2020.

\bibitem{fumero2011rim}
Francisco Fumero, Silvia Alay{\'o}n, Jos{\'e}~L Sanchez, Jose Sigut, and M~Gonzalez-Hernandez,
\newblock ``Rim-one: An open retinal image database for optic nerve evaluation,''
\newblock in {\em 2011 24th international symposium on computer-based medical systems (CBMS)}. IEEE, 2011, pp. 1--6.

\bibitem{sivaswamy2015comprehensive}
Jayanthi Sivaswamy, S~Krishnadas, Arunava Chakravarty, G~Joshi, A~Syed Tabish, et~al.,
\newblock ``A comprehensive retinal image dataset for the assessment of glaucoma from the optic nerve head analysis,''
\newblock {\em JSM Biomedical Imaging Data Papers}, vol. 2, no. 1, pp. 1004, 2015.

\end{thebibliography}

\end{document}